\documentclass[runningheads]{llncs}

\usepackage{epsfig}
\usepackage{graphicx}
\usepackage{amsmath}
\usepackage{amssymb}
\usepackage{caption}
\usepackage{booktabs}
\usepackage{multirow}
\usepackage[misc,geometry]{ifsym}
\usepackage{siunitx}
\usepackage{color}
\usepackage{xcolor}
\definecolor{OliveGreen}{rgb}{0,0.6,0}
\definecolor{SoftRed}{rgb}{1,0.2,0.2}
\usepackage[pagebackref]{hyperref}   
\hypersetup{
     colorlinks=true,
     linkcolor=cyan,
     filecolor=cyan,
     citecolor = OliveGreen,      
     urlcolor=cyan,
     }
\usepackage[hyperpageref]{backref}

\begin{document}

\title{BoundaryNet: An Attentive Deep Network with Fast Marching Distance Maps for Semi-automatic Layout Annotation}
\titlerunning{BoundaryNet}
%
\author{Abhishek Trivedi\orcidID{0000-0002-6763-4716} \and
Ravi Kiran Sarvadevabhatla (\Letter)\orcidID{0000-0003-4134-1154} }

\authorrunning{Trivedi and Sarvadevabhatla}
%
\institute{Centre for Visual Information Technology (CVIT)\\
International Institute of Information Technology, Hyderabad -- 500032, INDIA\\
\url{http://ihdia.iiit.ac.in/BoundaryNet/} \\
\email{\{abhishek.trivedi@research.,ravi.kiran@\}iiit.ac.in}}
\maketitle              
\begin{abstract}
Precise boundary annotations of image regions can be crucial for downstream applications which rely on region-class semantics. Some document collections contain densely laid out, highly irregular and overlapping multi-class region instances with large range in aspect ratio. Fully automatic boundary estimation approaches tend to be data intensive, cannot handle variable-sized images and produce sub-optimal results for aforementioned images. To address these issues, we propose BoundaryNet, a novel resizing-free approach for high-precision semi-automatic layout annotation. The variable-sized user selected region of interest is first processed by an attention-guided skip network. The network optimization is guided via Fast Marching distance maps to obtain a good quality initial boundary estimate and an associated feature representation. These outputs are processed by a Residual Graph Convolution Network optimized using Hausdorff loss to obtain the final region boundary. Results on a challenging image manuscript dataset demonstrate that BoundaryNet outperforms strong baselines and produces high-quality semantic region boundaries. Qualitatively, our approach generalizes across multiple document image datasets containing different script systems and layouts, all without additional fine-tuning. We integrate BoundaryNet into a document annotation system and show that it provides high annotation throughput compared to manual and fully automatic alternatives. 
\end{abstract}

\section{Introduction}

Document images exhibit incredible diversity in terms of language~\cite{yalniz2011fast,slimane2009database,clausner2019icdar2019}, content modality (printed~\cite{harley2015icdar,shahab2010open}, handwritten~\cite{simistira2016diva,8977999,kassis2017vml,897367}), writing surfaces (paper, parchment~\cite{DBLP:conf/icdar/PalTW13}, palm-leaf~\cite{8978062,kesiman2018benchmarking}), semantic elements such as text, tables, photos, graphics~\cite{clausner2019icdar2019,shahab2010open,8977963} and other such attributes. Within this variety, handwritten and historical documents pose the toughest challenges for tasks such as Optical Character Recognition (OCR) and document layout parsing. 

\begin{figure*}[!t]
   \centering
   \includegraphics[width=\textwidth]{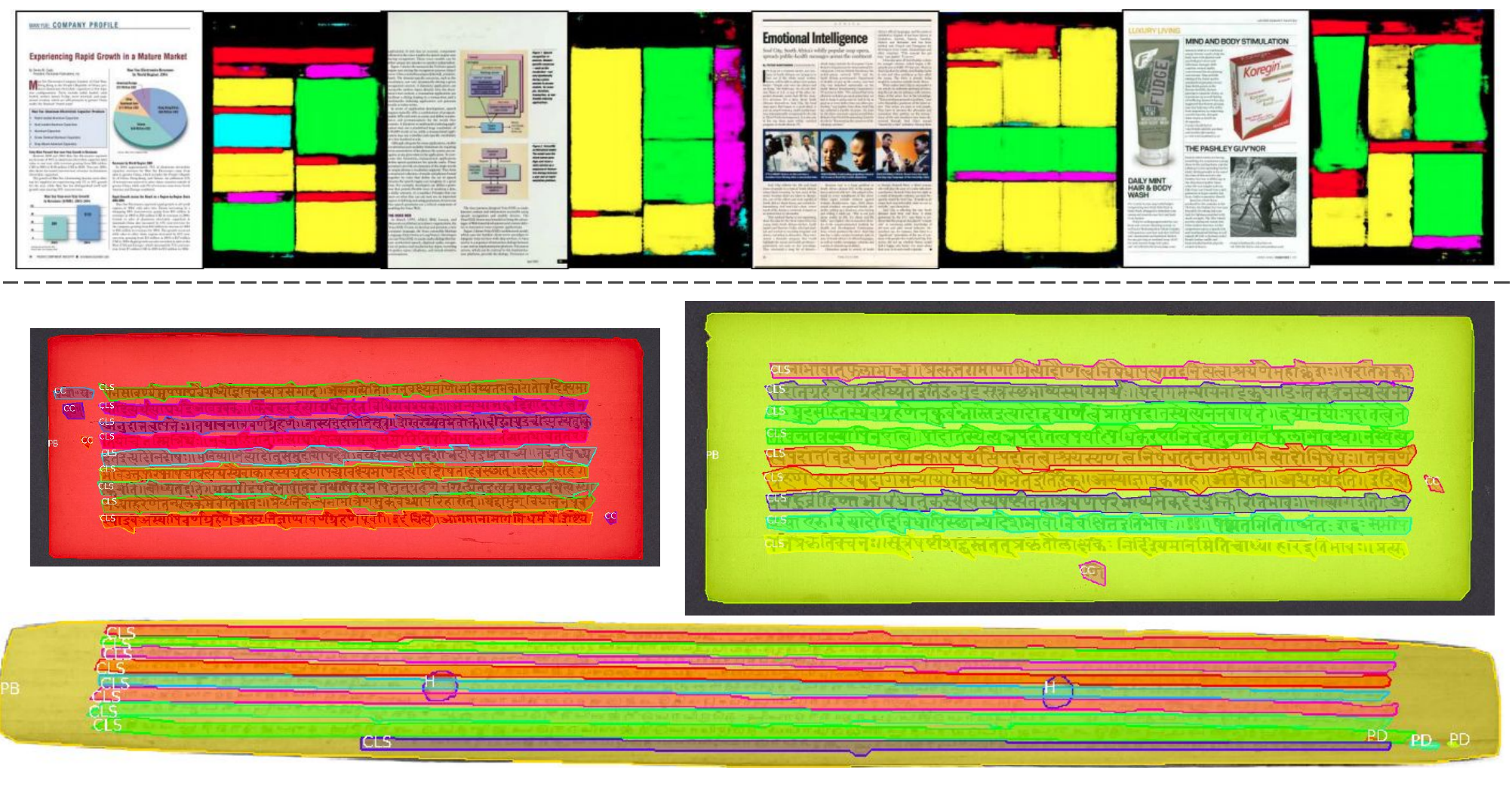}
   \caption{Compare the contours of semantic region instances for printed documents (top)~\cite{yang2017learning} and historical document images (bottom). The latter are very diverse, often found damaged, contain densely laid out overlapping region instances (lines, holes) with large range in aspect ratios and high local curvature. These factors pose a challenge for region annotation.}
   \label{fig:introfig}
\end{figure*}

In this work, we focus on historical documents. These documents form an important part of world's literary and cultural heritage. The mechanised process of machine printing imparts structure to modern-era paper documents. In contrast, historical document images are typically handwritten, unstructured and often contain information in dense, non-standard layouts (Fig.~\ref{fig:introfig}). Given the large diversity in language, script and non-textual elements in these documents, accurate spatial layout parsing can assist performance for other document-based tasks such as word-spotting~\cite{2019arXiv190709404L}, optical character recognition (OCR), style or content-based retrieval~\cite{Wiggers2019DeepLA,10.1007/978-981-13-9714-1_24}. Despite the challenges posed by such images, a number of deep-learning based approaches have been proposed for fully automatic layout parsing~\cite{simistira2016diva,8978062,Breuel2017RobustSP,ma2020segmentation}. However, a fundamental trade off exists between global processing and localized, compact nature of semantic document regions. For this reason, fully automatic approaches for documents with densely spaced, highly warped regions often exhibit false negatives or imprecise region boundaries. In practice, correction of predicted boundaries can be more burdensome than manual annotation itself. 

Therefore, we propose an efficient semi-automatic approach for parsing images with dense, highly irregular layouts. The user selected bounding-box enclosing the region of interest serves as a weak supervisory input. Our proposed deep neural architecture, BoundaryNet, processes this input to generate precise region contours which require minimal to no manual post-processing. 

Numerous approaches exist for weakly supervised bounding-box based semantic parsing of scene objects~\cite{bonechi2019coco_ts,song2019box}. However, the spatial dimensions and aspect ratios of semantic regions in these datasets are less extreme compared to ones found in handwritten documents (Fig.~\ref{fig:introfig}). More recently, a number of approaches model the annotation task as an active contour problem by regressing boundary points on the region's contour~\cite{acuna2018efficient,ling2019fast,marcos2018learning,8864970,acdrnet}. However, the degree of curvature for document region contours tends to be larger compared to regular object datasets. The image content and associated boundaries are also distorted by the standard practice of resizing the image to a common height and width. For these reasons, existing approaches empirically tend to produce imprecise contours, especially for regions with high warp, extreme aspect ratio and multiple curvature points (as we shall see).

To address these shortcomings, we propose a two-stage approach (Sec.~\ref{sec:BoundaryNet}). In the first stage, the variable-sized input image is processed by an attention-based fully convolutional network to obtain a region mask (Sec.~\ref{sec:encoder}). The region mask is morphologically processed to obtain an initial set of boundary points (Sec.~\ref{sec:contourize}). In the second stage, these boundary points are iteratively refined using a Residual Graph Convolutional Network to generate the final semantic region contour (Sec.~\ref{sec:gcn}). As we shall show, our design choices result in a high-performing system for accurate document region annotation.

Qualitatively and quantitatively, BoundaryNet outperforms a number of strong baselines for the task of accurate boundary generation (Sec.~\ref{sec:experiments}). BoundaryNet handles variable-sized images without resizing, in real-time, and generalizes across document image datasets with diverse languages, script systems and dense, overlapping region layouts (Sec.~\ref{sec:experiments}). Via end-to-end timing analysis, we showcase BoundaryNet's superior annotation throughput compared to manual and fully-automatic approaches (Sec.~\ref{sec:timing}). 

Source code, pre-trained models and associated documentation are available at  \url{http://ihdia.iiit.ac.in/BoundaryNet/}.

\section{Related Work}

Annotating spatial regions is typically conducted in three major modes -- manual, fully automatic and semi-automatic. The manual mode is obviously labor-intensive and motivates the existence of the other two modes. Fully automatic approaches fall under the task categories of semantic segmentation~\cite{fu2019stacked} and instance segmentation~\cite{he2017mask}. These approaches work reasonably well for printed~\cite{clausner2019icdar2019,8977963} and structured handwritten documents~\cite{Breuel2017RobustSP}, but have been relatively less successful for historical manuscripts and other highly unstructured documents containing distorted, high-curvature regions~\cite{kassis2017vml,8978062}. 

Given the challenges with fully automatic approaches, semi-automatic variants operate on the so-called `weak supervision' provided by human annotators. The weak supervision is typically provided as class label~\cite{tang2018regularized,8395163}, scribbles~\cite{can2018learning,7490105,7814032} or bounding box~\cite{bonechi2019coco_ts} for the region of interest with the objective of predicting the underlying region's spatial support. This process is repeated for all image regions relevant to the annotation task. In our case, we assume box-based weak supervision. Among box-based weakly supervised approaches, spatial support is typically predicted as a 2-D mask~\cite{bonechi2019coco_ts} or a boundary contour~\cite{acuna2018efficient,ling2019fast,marcos2018learning,8864970,acdrnet}. 

Contour-based approaches generally outperform mask-based counterparts and provide the flexibility of semi-automatic contour editing~\cite{acuna2018efficient,8864970,marcos2018learning,acdrnet,ling2019fast}. We employ a contour-based approach. However, unlike existing approaches, (i) BoundaryNet efficiently processes variable-sized images without need for resizing (ii) Boundary points are adaptively initialized from an inferred estimate of region mask instead of a fixed geometrical shape (iii) BoundaryNet utilizes skip connection based attentional guidance and boundary-aware distance maps to semantically guide region mask production (iv) BoundaryNet also produces region class label reducing post-processing annotation efforts. Broadly, our choices help deal with extreme aspect ratios and highly distorted region boundaries typically encountered in irregularly structured images.     

\section{BoundaryNet}
\label{sec:BoundaryNet}

\begin{figure*}[!ht]
   \centering
   \includegraphics[width=\textwidth]{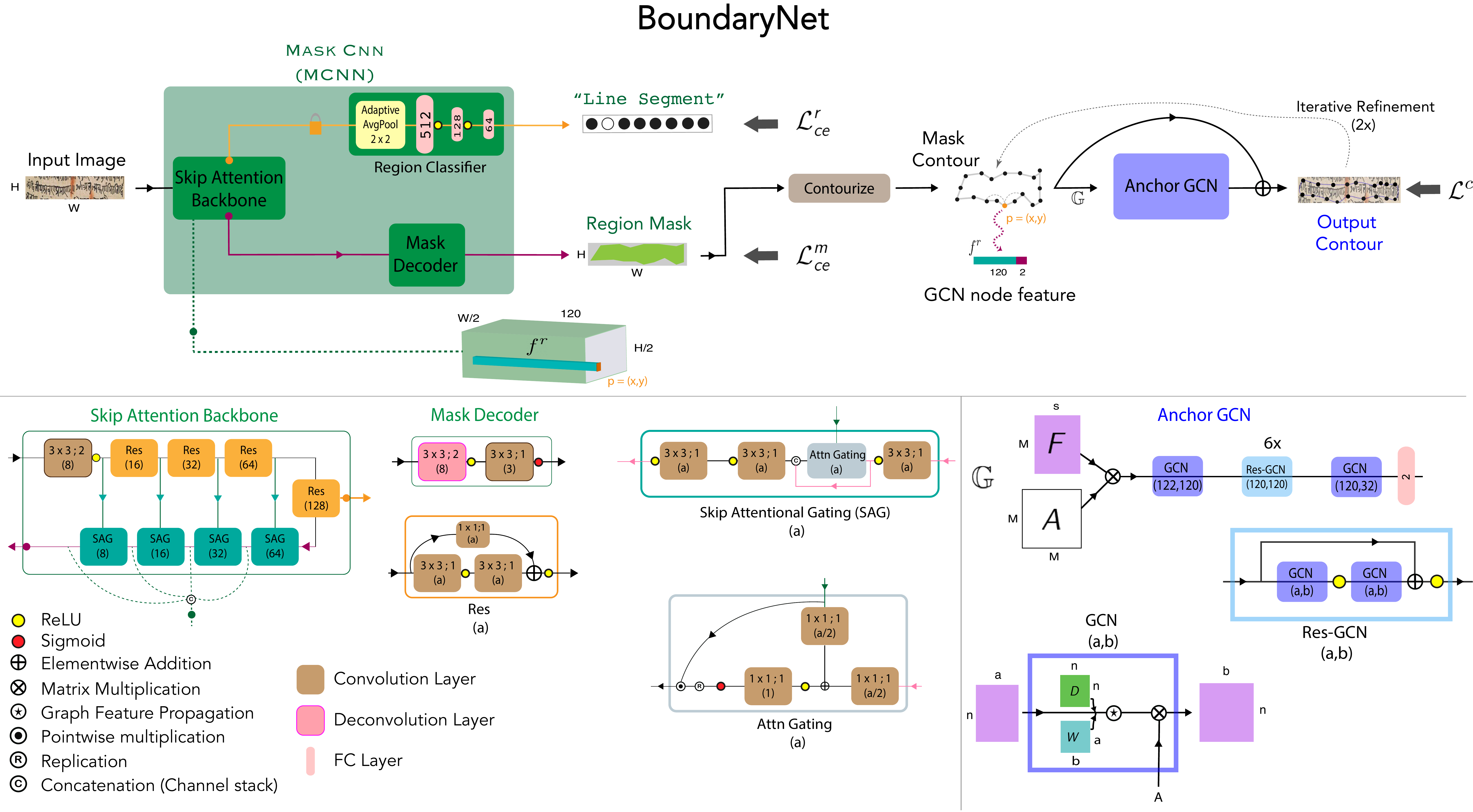}
   \caption{The architecture of BoundaryNet (top) and various sub-components (bottom). The variable-sized $H \times W$ input image is processed by Mask-CNN (MCNN) which predicts a region mask estimate and an associated region class (Sec. \ref{sec:encoder}). The mask's boundary is determined using a contourization procedure (light brown) applied on the estimate from MCNN. $M$ boundary points are sampled on the boundary (Sec. \ref{sec:contourize}). A graph is constructed with the points as nodes and edge connectivity defined by $\leqslant k$-hop neighborhoods of each point. The spatial coordinates of a boundary point location $p=(x,y)$ and corresponding backbone skip attention features from MCNN $f^r$ are used as node features for the boundary point. The feature-augmented contour graph $\mathbb{G}=(F,A)$ is iteratively processed by Anchor GCN (Sec.~\ref{sec:gcn}) to obtain the final output contour points defining the region boundary. Note that all filters in MCNN have a $3 \times 3$ spatial extent. The orange lock symbol on region classifier branch indicates that it is trained standalone, i.e. using pre-trained MCNN features.}
   \label{fig:BoundaryNet}
\end{figure*}

\noindent \textbf{Overview:} Given the input bounding box, our objective is to obtain a precise contour of the enclosed semantic region (e.g. text line, picture, binding hole). BoundaryNet's processing pipeline consists of three stages -- see Fig.~\ref{fig:BoundaryNet}. In the first stage, the bounding box image is processed by a Mask-CNN (MCNN) to obtain a good quality estimate of the underlying region's spatial mask (Sec.~\ref{sec:encoder}). Morphological and computational geometric procedures are used to sample contour points along the mask boundary (Sec.~\ref{sec:contourize}). A graph is constructed with contour points as nodes and edge connectivity defined by local neighborhoods of each contour point. The intermediate skip attention features from MCNN and contour point location are used to construct feature representations for each graph node. Finally, the feature-augmented contour graph is processed by a Graph Convolutional Network (Anchor GCN - Sec.~\ref{sec:gcn}) iteratively to obtain final set of contour points which define the predicted region boundary. 

Semantic regions in documents are often characterized by extreme aspect ratio variations across region classes and uneven spatial distortion. In this context, it is important to note that BoundaryNet processes the input as-is without any resizing to arbitrarily fixed dimensions. This helps preserve crucial appearance detail. Next, we describe the components of BoundaryNet.

\subsection{Mask-CNN (MCNN)}
\label{sec:encoder}

As the first step, the input image is processed by a backbone network (`Skip Attention Backbone' in Fig.~\ref{fig:BoundaryNet}). The backbone has U-Net style long-range skip connections with the important distinction that no spatial downsampling or upsampling is involved. This is done to preserve crucial boundary information. In the first part of the backbone, a series of residual blocks are used to obtain progressively refined feature representations (orange blocks). The second part of the backbone contains another series of blocks we refer to as Skip Attentional Guidance (SAG). Each SAG block produces increasingly compressed (channel-wise) feature representations of its input. To accomplish this feat without losing crucial low-level feature information, the output from immediate earlier SAG block is fused with skip features originating from a lower-level residual block layer (refer to `Skip Attention Backbone' and its internal module diagrams in Fig.~\ref{fig:BoundaryNet}). This fusion is modulated via an attention mechanism (gray `Attn Gating' block)~\cite{attentionunet}.

The final set of features generated by skip-connection based attentional guidance (magenta) are provided to the `Mask Decoder' network which outputs a region mask binary map. In addition, features from the last residual block (\texttt{Res-128}) are fed to  `Region Classifier' sub-network which predicts the associated region class. Since input regions have varying spatial dimensions, we use adaptive average pooling~\cite{he2015spatial} to ensure a fixed-dimensional fully connected layer output (see `Region Classifier' in Fig.~\ref{fig:BoundaryNet}).

The input image is processed by an initial convolutional block with stride $2$ filters before the resulting features are relayed to the backbone residual blocks. The spatial dimensions are restored via a transpose convolution upsampling within `Mask Decoder' sub-network. These choices help keep the feature representations compact while minimizing the effect of downsampling.  

\subsection{Contourization}
\label{sec:contourize}

\begin{figure*}[!t]
  \centering
  \includegraphics[width=\textwidth]{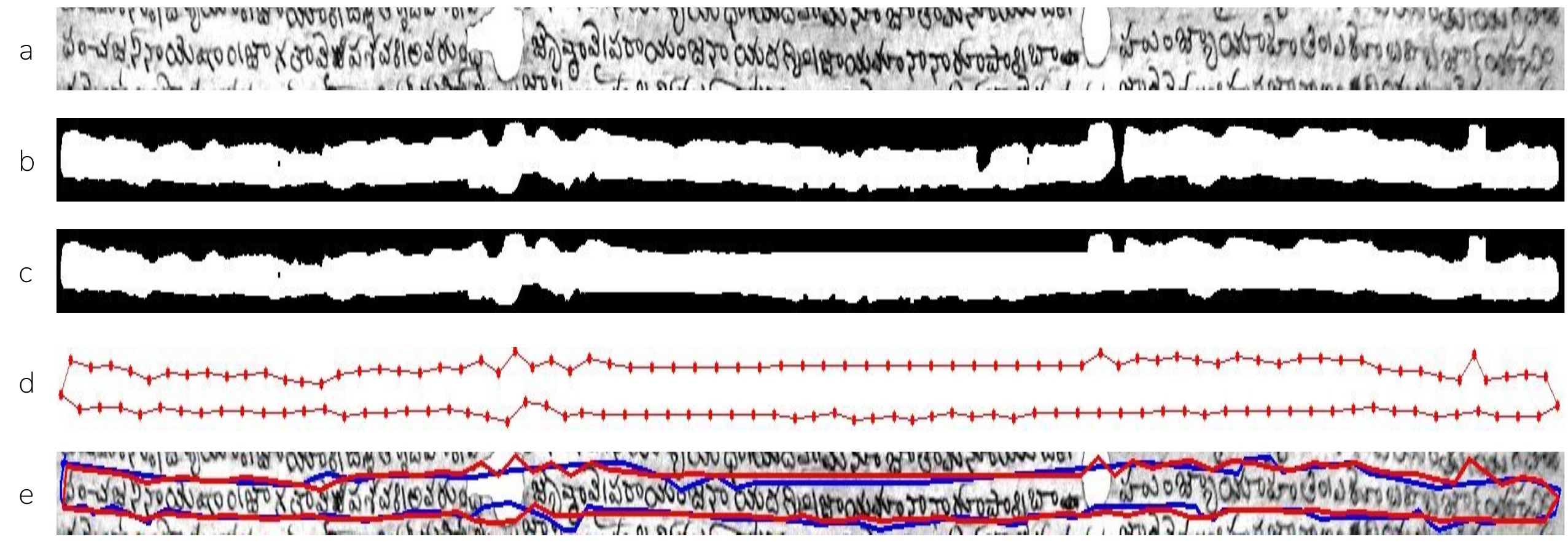}
  \caption{Contourization (Sec.~\ref{sec:contourize}): a - input image, b - thresholded initial estimate from MCNN, c - after area-based thresholding and joining centroids of largest connected components by an adaptive $m=\frac{H}{7}$ pixel-thick line where $H$ is the height of the input image, d - after morphological closing, contour extraction,  b-spline fitting and uniform point sampling, e - estimated contour (red) and ground-truth (blue) overlaid on input image.}
  \label{fig:contourization}
\end{figure*}

The pixel predictions in the output from `Mask Decoder' branch are thresholded to obtain an initial estimate of the region mask. The result is morphologically processed, followed by the extraction of mask contour. A b-spline representation of mask contour is further computed to obtain a smoother representation. $M$ locations are uniformly sampled along the b-spline contour curve to obtain the initial set of region boundary points. Figure~\ref{fig:contourization} illustrates the various steps.

An advantage of the above procedure is that the set of mask-based boundary points serves as a reasonably accurate estimate of the target boundary. Therefore, it lowers the workload for the subsequent GCN stage which can focus on refining the boundary estimate. 

\subsection{Anchor GCN}
\label{sec:gcn}

The positional and appearance-based features of boundary points from the contourization stage (Sec.~\ref{sec:contourize}) are used to progressively refine the region's boundary estimate. For this, the boundary points are first assembled into a contour graph. The graph's connectivity is defined by $\leqslant \!\!k$-hop neighbors for each contour point node. The node's $s$-dimensional feature representation is comprised of (i) the contour point 2-D coordinates $p=(x,y)$ (ii)  corresponding skip attention features from MCNN $f^r$ - refer to `GCN node feature' in Fig.~\ref{fig:BoundaryNet} for a visual illustration.

The contour graph is represented in terms of two matrices - feature matrix $F$ and adjacency matrix $A$~\cite{kipf2016variational,zhang2018graph}. $F$ is a $\mathsf{M} \times s$ matrix where each row corresponds to the $s$-dimensional boundary point feature representation described previously. The $\mathsf{M} \times \mathsf{M}$ binary matrix $A$ encodes the $\leqslant \!\!k$-hop connectivity for each boundary point. Thus, we obtain the contour graph representation $\mathbb{G} = (F,A)$ (denoted `Residual Graph' at bottom-right of Fig.~\ref{fig:BoundaryNet}). We briefly summarize GCNs next.

\smallbreak

\noindent \textbf{Graph Convolutional Network (GCN):} A GCN takes a graph $\mathbb{G}$ as input and computes hierarchical feature representations at each node in the graph while retaining the original connectivity structure. The feature representation at the ($i+1$)-th layer of the GCN is defined as $H_{i+1} = f(H_i, A)$ where $H_i$ is a $\mathsf{p} \times F_i$  matrix whose $j$-th row contains the $i$-th layer's feature representation for node indexed by $j$ ($1 \leqslant j \leqslant \mathsf{N}$). $f$ (the so-called propagation rule) determines the manner in which node features of previous layer are aggregated to obtain current layer's feature representation. We use the following propagation rule~\cite{kipf2017semi}:

\begin{align}
    f(H_i,A) = \sigma({D}^{\frac{-1}{2}} \widetilde{A} {D}^{\frac{-1}{2}} H_i W_i) 
\end{align}

\noindent where $\widetilde{A} = A + I$ represents the adjacency matrix modified to include self-loops, ${D}$ is a diagonal node-degree matrix (i.e. ${D}_{jj} = \sum_m  \widetilde{A}_{jm}$) and $W_i$ are the trainable weights for $i$-th layer. $\sigma$ represents a non-linear activation function (ReLU in our case). Also, $H_0 = F$ (input feature matrix). 

\smallbreak

\textbf{Res-GCN:} The residual variant of GCN operates via an appropriate `residual' modification to the GCN layer's feature representation and is defined as $H_{i+1} = f(H_i, A) + H_i$.

\smallbreak
     
\noindent The input contour graph features are processed by a series of Res-GCN blocks~\cite{li2019deepgcns} sandwiched between two GCN blocks. The Anchor GCN module culminates in a 2-dimensional fully connected layer whose output constitutes per-point displacements of the input boundary locations. To obtain the final boundary, we perform iterative refinement of predicted contour until the net displacements are negligibly small by re-using GCN's prediction for the starting estimate at each iteration~\cite{ling2019fast}. 

\subsection{Training and Inference}
\label{sec:training}

\noindent We train BoundaryNet in three phases. 

\noindent \textbf{First Phase:} In this phase, we aim to obtain a good quality estimate of the boundary contour from MCNN. For this, the binary prediction from MCNN is optimized using  per-pixel class-weighted binary focal loss~\cite{lin2017focal}:

\begin{align}
{l}_{BFL} = {\alpha}_{c} y {(1-p)}^{\gamma} \cdot \log p + (1-y) {p}^{\gamma} \cdot \log (1-p)
\label{eqn:bfl}
\end{align}

\noindent where $y \in \{0,1\}$ is ground-truth label, $p$ is the corresponding pixel-level prediction, $\alpha_{c}$ = $N_{b}/N_{f}$ is the ratio of background to foreground (region mask) pixel counts and $\gamma$ is the so-called focusing hyperparameter in focal loss. The class-weighting ensures balanced optimization for background and foreground, indirectly aiding contour estimation. The focal loss encourages the optimization to focus on the harder-to-classify pixels.

\begin{figure*}[!t]
   \centering
   \includegraphics[width=0.945\textwidth]{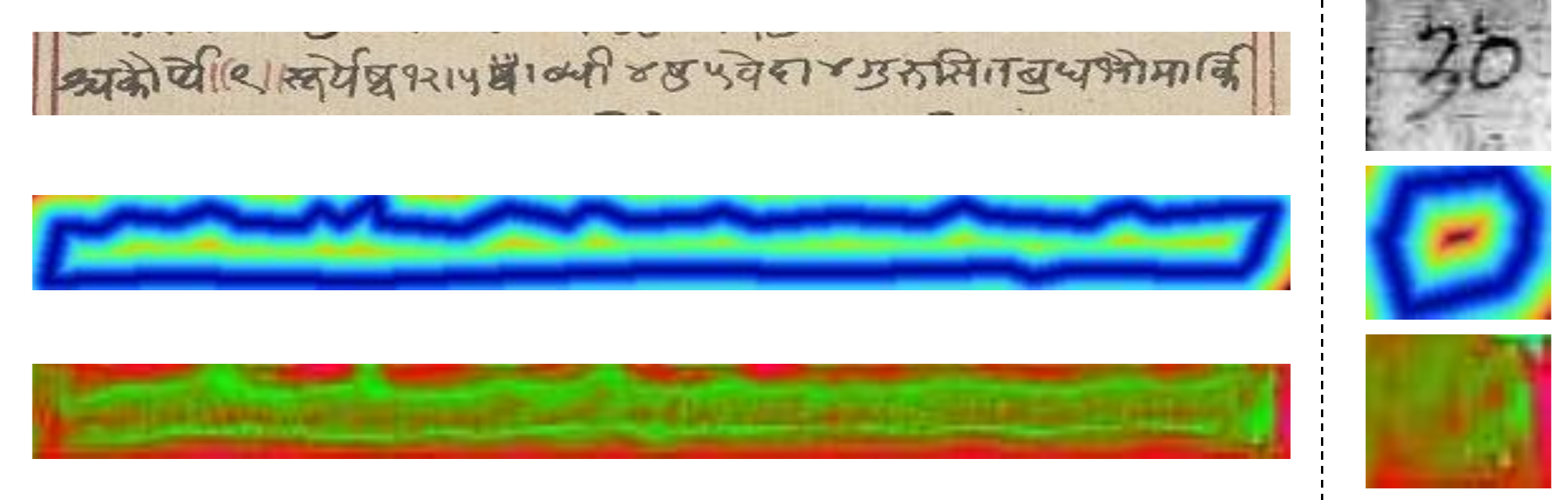}
   \caption{Some examples of region image (top) and color-coded versions of fast marching distance map (middle) and the attention map computed by final SAG block of BoundaryNet (bottom). The relatively larger values at the crucial boundary portions can be clearly seen in the attention map.}
   \label{fig:fmd}
\end{figure*}

To boost the precision of estimate in a more boundary-aware manner, we first construct a distance map using a variant of the Fast Marching method~\cite{sethian1996fast}. The procedure assigns a distance of $0$ to  boundary pixels and progressively higher values to pixels based on contours generated by iterative erosion and dilation of ground-truth region mask (see Figure~\ref{fig:fmd}. The distance map is inverted by subtracting each entry from the global maximum within the map. Thus, the highest weights are assigned to boundary pixels, with the next highest set of values for pixels immediately adjacent to the boundary. The inverted distance map is then normalized ($[0,1]$) to obtain the final map $\Psi$. The class-weighted binary focal loss spatial map $\mathcal{L}_{BFL}$ is constituted from per-pixel losses  ${l}_{BFL}$ (Eq.~\ref{eqn:bfl}) and further weighted by $\Psi$ as follows:

\begin{align}
 \mathcal{L}_{FM} = (1+\Psi) \odot \mathcal{L}_{BFL}
 \label{eqn:fm}
\end{align}

\noindent where $\odot$ stands for the Hadamard product. The above formulation is preferred to solely weighting $\mathcal{L}_{BFL}$ with $\Psi$ to mitigate the vanishing gradient issue.

\smallbreak

\begin{table*}[!t]
\captionof{table}{Train, Validation and Test split for different region types.}
\centering
\resizebox{\textwidth}{!}{%
 \centering 
 \begin{tabular}{l|c|cccccccc}
 \toprule 
 Split & \texttt{Total} & Hole & Line Segment & Degradation & Character & Picture & Decorator & Library Marker & Boundary Line \\
 \midrule
 Train   &   $6491$    &  $422$ & $3535$  & $1502$  & $507$ & $81$  & $48$ & $116$  & $280$  \\
 Validation   &     $793$ &  $37$ & $467$  & $142$  & $75$ & $5$  & $12$ & $18$  & $37$  \\
 Test &     $1633$      &  $106$ & $836$  & $465$     &  $113$ & $9$  & $5$ & $31$  & $68$  \\
 \bottomrule
 \end{tabular}
 }
\label{tab:trainvaltest} 
\end{table*}

\noindent \textbf{Second Phase:} In this phase of training, MCNN's weights are frozen and the estimate of region mask is obtained as described previously (Sec.~\ref{sec:encoder}). The contour graph constructed from points on region mask boundary (Sec.~\ref{sec:contourize}) is fed to Anchor GCN. The output nodes from Anchor GCN are interpolated $10\times$ through grid sampling. This ensures maximum optimal shifts towards ground-truth contour for original graph nodes and avoids graph distortion. 

Let $\mathcal{G}$ be the set of points (x-y locations) in ground-truth contour and $\mathcal{B}$, the point set predicted by Anchor GCN. Let $E_1$ be the list of minimum Euclidean distances calculated per ground-truth point $g_i \in \mathcal{G}$ to a point in $\mathcal{B}$, i.e. $e_i = \displaystyle \min\limits_{j}\ \parallel g_i - b_j \parallel, e_i \in E_1, b_j \in \mathcal{B}$. Let $E_2$ be a similar list obtained by flipping the roles of ground-truth and predicted point sets. The Hausdorff Distance loss~\cite{ribera2019} for optimizing Anchor GCN is defined as: 

\begin{align}
    {L}_{C}(E_1,E_2) = 0.5 ( \displaystyle \sum_i e_i + \displaystyle \sum_j e_j )
\label{eqn:ft}
\end{align}
\noindent where $e_j \in E_2$.

\smallbreak

\noindent \textbf{Third Phase:} In this phase, we jointly fine-tune the parameters of both MCNN and Anchor GCN in an end-to-end manner. The final optimization is performed by minimizing ${L}_{FT}$ loss defined as: ${L}_{FT} = {L}_{C} + \lambda \; \mathcal{L}_{FM}$. As we shall see, the end-to-end optimization is crucial for improved performance  (Table~\ref{table:ablations}).

\smallbreak

The region classification sub-branch is optimized using categorical cross-entropy loss ($\mathcal{L}_{CE}^r$) after all the phases mentioned above. During this process, the backbone is considered as a pre-trained feature extractor, i.e. backpropagation is not performed on MCNN backbone's weights.

\subsection{Implementation Details}
\label{sec:impl}

\noindent \textbf{MCNN:} The implementation details of MCNN can be found in Fig.~\ref{fig:BoundaryNet}. The input $H \times W \times 3$ RGB image is processed by MCNN to generate a corresponding $H \times W $ region mask representation (magenta branch in Fig.~\ref{fig:BoundaryNet}) and a region class prediction (orange branch) determined from the final $8$-way softmax layer of the branch. In addition, the outputs from the SAG blocks are concatenated and the resulting $\frac{H}{2} \times \frac{W}{2} \times 120$ output (shown at the end of dotted green line in Fig.~\ref{fig:BoundaryNet}) is used to determine the node feature representations $f^r$ used in the downstream Anchor GCN module. 

For MCNN training, the focal loss (Eq.~\ref{eqn:bfl}) is disabled at the beginning, i.e. $\gamma = 0$. The batch size is set to $1$ with an initial learning rate of $3e^{-5}$. A customized variant of Stochastic Gradient Descent with Restart~\cite{loshchilov2016sgdr} is conducted. Two fresh restarts are performed by increasing learning rate $5\times$ for $3$ epochs and dropping it back to counter potential loss saturation. The focal loss is invoked with $\gamma = 2$ when $\mathcal{L}_{FM}$ (Eq.~\ref{eqn:fm}) starts to plateau. At this stage, the learning rate is set to decay by $0.5$ every $7$ epochs.

\smallbreak

\noindent \textbf{Contourization:} The region formed by pixels labelled as region interior in MCNN's output is morphologically closed using a $3 \times 3$ disk structuring element. Major regions are extracted using area-based thresholding. The final region interior mask is obtained by connecting all the major sub-regions through their centroids. A b-spline curve is fit to the boundary of the resulting region and $M=200$ boundary points are uniformly sampled along the curve - this process is depicted in Fig.~\ref{fig:contourization}. 

\smallbreak

\noindent \textbf{Anchor GCN:} Each boundary point's $122$-dimensional node feature is obtained by concatenating the $120$-dimensional feature column ($f^r$ in Fig.~\ref{fig:BoundaryNet}) and the point's 2-D coordinates $p=(x,y)$ (normalized to a $[0,1] \times [0,1]$ grid). Each contour point is connected to its $20$ nearest sequential neighbors in the contour graph, ten on each side along the contour (see `Mask Contour' in Fig.~\ref{fig:BoundaryNet}), i.e. maximum hop factor $k=10$. The graph representation is processed by two GCN and six residual GCN layers (see `Residual GCN' in Fig.~\ref{fig:BoundaryNet} for architectural details). The resulting features are processed by a fully connected layer to produce 2-D residuals for each of the boundary points. The iterative refinement of  boundary points is performed two times. During training, the batch size is set to $1$ with a learning rate of $1e^{-3}$.

\smallbreak

\noindent \textbf{End-to-end framework:} For joint optimization, the batch size set to $1$ with a relatively lower learning rate of $1e^{-5}$. Weighting coefficient $\lambda$ (in Eq.~\ref{eqn:ft}) is set to $200$. 

Throughout the training phases and for loss computation, the predicted points and ground-truth are scaled to a unit normalized ($[0,1] \times [0,1]$) canvas. Also, to ensure uniform coverage of all region classes, we perform class-frequency based mini-batch resampling and utilize the resultant sequences for all phases of training. 

\begin{table*}[!t]
\captionof{table}{
Region-wise average and overall Hausdorff Distance (HD) for different baselines and BoundaryNet on Indiscapes dataset.}
\resizebox{\textwidth}{!}{%
\centering
 \centering 
 \begin{tabular}{r|c|cccccccc}
 \toprule 
        &   \texttt{HD} $\downarrow$           & Hole & Line Segment & Degradation & Character & Picture & Decorator & Library Marker & Boundary Line \\
 \midrule
\textbf{BoundaryNet}      &     ${\mathbf{17.33}}$  &  $6.95$ & $20.37$  & $10.15$  & $7.58$ & $51.58$  & $20.17$ & $16.42$  & $5.45$  \\
Polygon-RNN++\cite{acuna2018efficient} &     $30.06$      &  $5.59$ & $66.03$  & $7.74$ &  $5.11$ & $ 105.99$  & $25.11$ & $9.97$  & $15.01$  \\
Curve-GCN\cite{ling2019fast}  &  $39.87$     &  $8.62$ & $142.46$  & $14.55$  & $10.25$ & $68.64$  & $32.11$ & $19.51$  & $22.85$  \\
DACN\cite{8864970}   &    $41.21$   &  $8.48$ & $105.61$  & $14.10$  & $11.42$ & $91.18$  & $26.55$ & $22.24$  & $50.16$  \\
DSAC\cite{marcos2018learning}   &   $54.06$    &  $14.34$ & $237.46$    & $10.40$ & $8.27$  & $65.81$ & $39.36$  & $23.34$ & $33.53$ \\
 \bottomrule
 \end{tabular}
 }
\label{table:HD} 
\end{table*}

\section{Experimental Setup}
\label{sec:experiments}

\noindent \textbf{Performance Measure:} As performance measure, we use the Hausdorff Distance (HD)~\cite{hausdorff} between the predicted contour and its ground-truth counterpart (Sec.~\ref{sec:impl}). Note that smaller the HD, the better is the boundary prediction. The per-region HD is obtained as the average over the HD of associated region instances.

For all the models, we use performance on the validation set to determine the optimal hyperparameters and determine architectural choices. Subsequently, we optimize the models on the combined training and validation splits and conduct a one-time evaluation on the test split.

\noindent \textbf{Baselines:} To perform a comparative evaluation of BoundaryNet, we include multiple state-of-the-art semi-automatic annotation approaches - DSAC~\cite{marcos2018learning}, 
Polygon-RNN++~\cite{acuna2018efficient}, Curve-GCN~\cite{ling2019fast} and DACN~\cite{8864970}. These approaches exhibit impressive performance for annotating semantic regions in street-view dataset and for overhead satellite imagery. However, directly fine-tuning the baselines resulted in bad performance due to the relatively fewer annotation nodes regression and domain gap between document images and imagery (street-view, satellite) for which the baselines were designed. Therefore, we use the original approaches as a guideline and train modified versions of the baseline deep networks.

\begin{figure*}[!t]
   \centering
   \includegraphics[width=\textwidth]{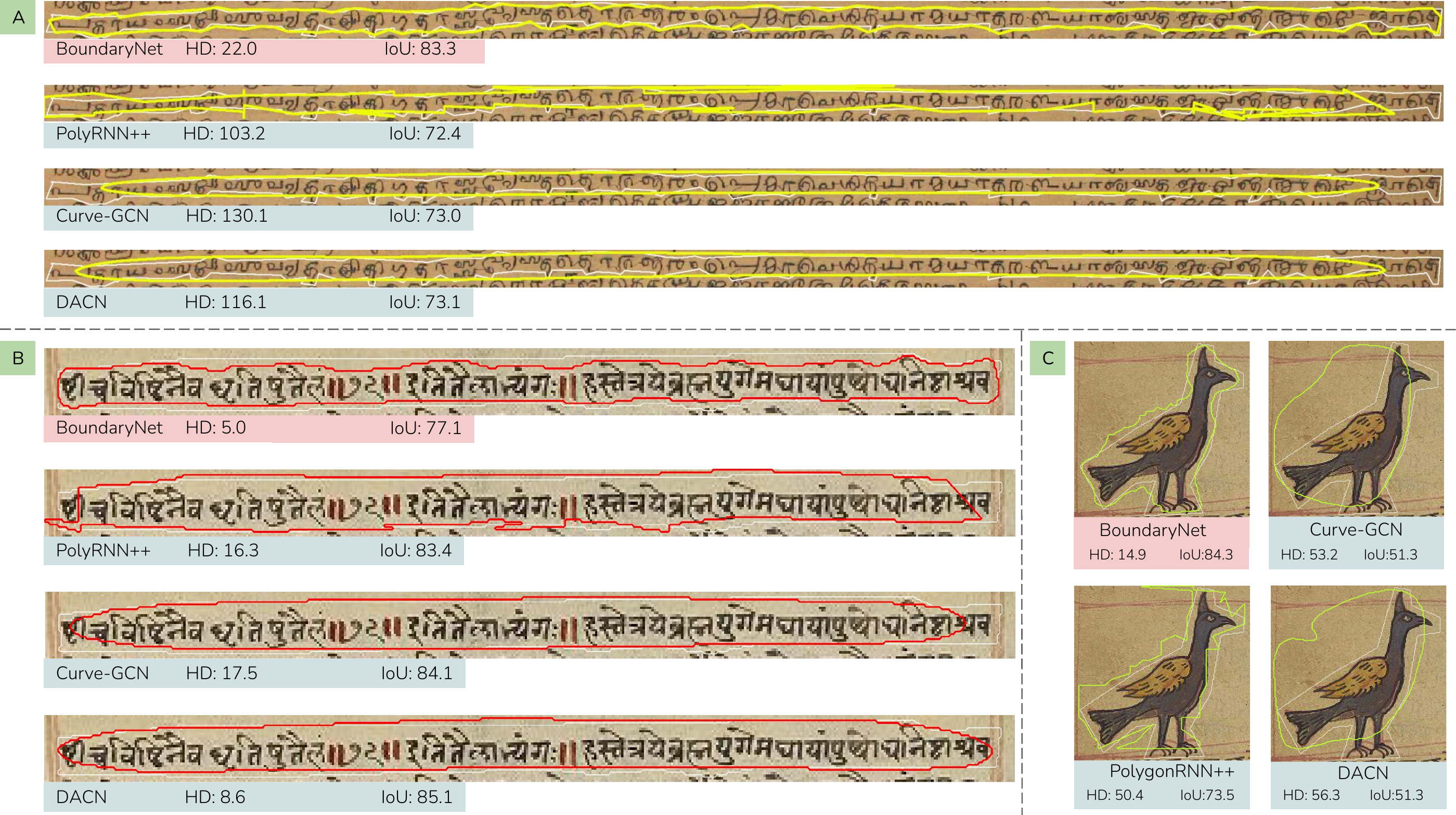}
   \caption{Qualitative comparison of BoundaryNet with baselines on sample test images from Indiscapes dataset. For each region, the ground-truth contour is outlined in white. The IoU score is also mentioned for reference (see Sec.~\ref{sec:results}).}
   \label{fig:baselines}
\end{figure*}

\noindent \textbf{Evaluation dataset:} For training and primary evaluation, we use Indiscapes~\cite{8978062}, a challenging historical document dataset of handwritten manuscript images. It contains $526$ diverse document images containing $9507$ regions spanning the following categories: Holes, Line Segments, Physical Degradation, Character Component, Picture, Decorator, Library Marker, Boundary Line, Page Boundary (omitted for our evaluation). Details of the training, validation and test splits can be viewed in Table~\ref{tab:trainvaltest}.  

\section{Results}
\label{sec:results}

\subsection{Indiscapes}

\begin{figure*}[!t]
   \centering
   \includegraphics[width=\textwidth]{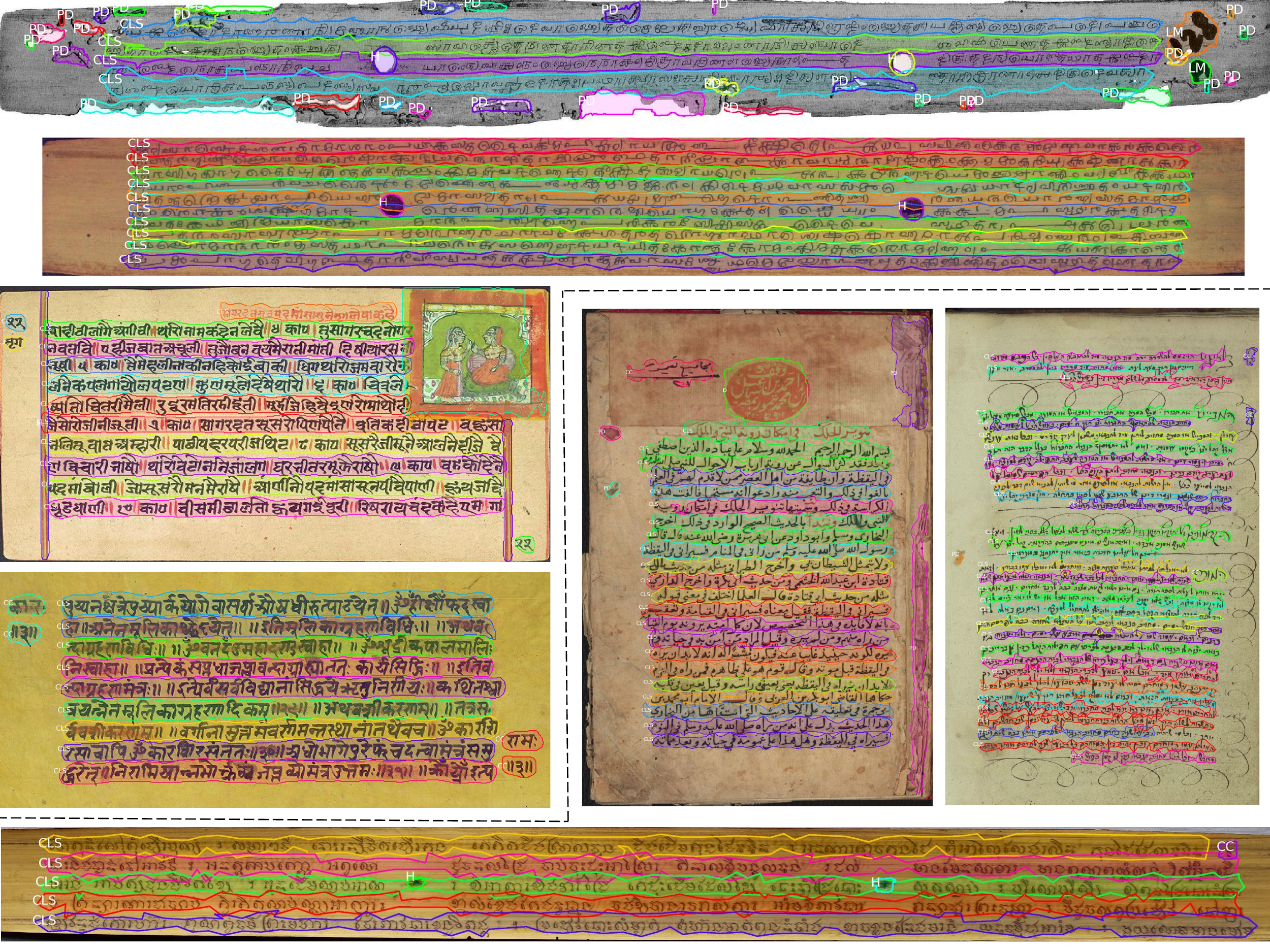}
   \caption{Semantic region boundaries predicted by BoundaryNet. The colors distinguish instances  -- they are \textit{not} region labels (written in shorthand alongside the regions). The dotted line separates Indiscapes dataset images (top) and those from other document collections (bottom). Note: BoundaryNet has been trained only on Indiscapes.}
   \label{fig:qualitative}
\end{figure*}

\noindent \textbf{Quantitative Baseline Comparison:} As Table~\ref{table:HD} shows, BoundaryNet outperforms other baselines by a significant margin in terms of overall Hausdorff Distance (HD). Considering that images in the test set have widths as large as $6800$ pixels, the results indicate a high degree of precision for obtained contours. The performance of BoundaryNet is slightly lower than the best on regions such as `Holes', `Library Marker' and `Degradation' due to the filtering induced by the GCN. However, notice that the performance for region present most frequently - `Line Segment' - is markedly better than other baselines. 

\smallbreak

\noindent \textbf{Qualitative Baseline Comparison:} The performance of BoundaryNet and top three  baseline performers for sample test images can be viewed in Figure~\ref{fig:baselines}. In addition to HD, we also mention the IoU score. As the results demonstrate, HD is more suited than IoU for standalone and comparative performance assessment of boundary precision. The reason is that IoU is an area-centric measure, suited for annotating objects in terms of their rigid edges (e.g. objects found in real-world scenery). As example B in Fig.~\ref{fig:baselines} shows, a boundary estimate which fails to enclose the semantic content of the region properly can still have a high IoU. In contrast, semantic regions found in documents, especially character lines, typically elicit annotations which aim to minimally enclose the region's semantic content in a less rigid manner. Therefore, a contour-centric measure which penalizes boundary deviations is more suitable. 

\smallbreak

\noindent \textbf{Qualitative Results (Image-level):} Examples of document images with BoundaryNet  predictions overlaid can be seen in Figure~\ref{fig:qualitative}. The images above the dotted line are from the Indiscapes dataset. The documents are characterized by dense layouts, degraded quality (first image), ultra wide character lines (second image). Despite this, BoundaryNet provides accurate annotation boundaries. Note that BoundaryNet also outputs region labels. This results in amortized time and labor savings for the annotator since region label need not be provided separately. Region Classifier performance can be seen in Figure~\ref{fig:TimeData} (left).

\smallbreak

\noindent \textbf{Performance on other document collections:} To determine its general utility, we used BoundaryNet for semi-automatic annotation of documents from other historical manuscript datasets (South-East Asian palm leaf manuscripts, Arabic and Hebrew documents). The results can be viewed in the images below the dotted line in Figure~\ref{fig:qualitative}. Despite not being trained on images from these collections, it can be seen that BoundaryNet provides accurate region annotations.

\smallbreak

\noindent \textbf{Ablations:} To determine the contribution of various architectural components, we examined lesioned variants of BoundaryNet for ablation analysis. The results can be viewed in Table~\ref{table:ablations}. As can be seen, the choices related to the MCNN's loss function, presence of error penalizing distance maps, number of points sampled on mask boundary estimate, spline interpolation, all impact performance in a significant manner.

\begin{table*}[!t]
\captionof{table}{Performance for ablative variants of BoundaryNet. The + refers to MCNN's output being fed to mentioned ablative variants of AGCN.}
\centering
\resizebox{\textwidth}{!}
{%
\begin{tabular}{lccc}
 \toprule 

 Component & Ablation type & Default Configuration in BoundaryNet & \texttt{HD} $\downarrow$  \\ 
 \midrule
 MCNN & Max Residual Channels=$64$ & Max Residual Channels=$128$                &  $21.77$     \\
 MCNN  & No Focal Loss & Focal Loss             &  $22.98$    \\
 MCNN  & No Fast Marching weights Penalization & Fast Marching weights Penalization             &  $23.96$    \\
 MCNN  & Normal skip connection, no attention gating & Skip connection with attention gating & $28.27$  \\
 MCNN  &   No AGCN         & AGCN & $19.17$  \\
 
+AGCN &  $\leqslant 5$-hop neighborhood &  $\leqslant 10$-hop neighborhood &  $19.26$    \\
+AGCN &  $\leqslant 15$-hop neighborhood &  $\leqslant 10$-hop neighborhood &  $20.82$    \\
+AGCN &  $1\times$ spline interpolation & $10\times$ interpolation &  $20.48$ \\
+AGCN &  $1$ iteration & $2$ iterations &  $19.31$ \\
+AGCN &  $100$ graph nodes & $200$ graph nodes &  $20.37$   \\
+AGCN &  $300$ graph nodes & $200$ graph nodes &  $19.98$  \\
+AGCN &  Node features: backbone only $f^r(x,y)$ & $f^r(x,y)$, $(x,y)$ & $20.50$ \\
+Fine-Tuning & No end-to-end finetuning & End to end finetuning & $18.79$ \\
\midrule
 BoundaryNet & --  & original & $\mathbf{17.33}$  \\
 \bottomrule
 \end{tabular}
 }

\label{table:ablations} 
\end{table*}

\subsection{Timing Analysis}
\label{sec:timing}

\begin{figure}[!t]
   \centering
   \includegraphics[width=\linewidth]{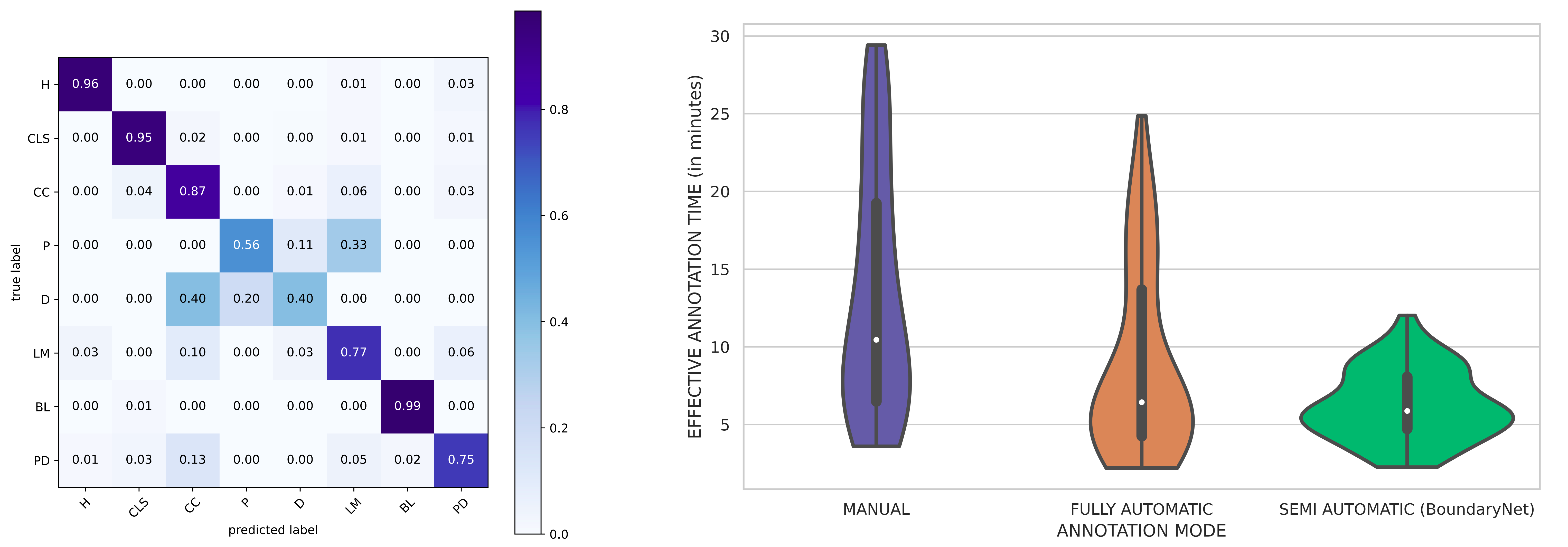}
   \caption{(left) Confusion Matrix from Region classifier branch, (right) Document-level end-to-end annotation duration distribution for various approaches depicted as a violin plot(the white dot represents mean duration - see Sec.~\ref{sec:timing}).}
   \label{fig:TimeData}
\end{figure}

To determine BoundaryNet utility in a practical setting, we obtained document-level annotations for test set images from Indiscapes dataset. The annotations for each image were sourced using an in-house document annotation system in three distinct modes: Manual Mode (hand-drawn contour generation and region labelling), Fully Automatic Mode (using an existing instance segmentation approach~\cite{8978062} with post-correction using the annotation system) and Semi-Automatic Mode (manual input of region bounding boxes which are subsequently sent to BoundaryNet, followed by post-correction). For each mode, we recorded the end-to-end annotation time at per-document level, including manual correction time. The distribution of annotation times for the three modes can be seen in Figure~\ref{fig:TimeData} (right). As can be seen, the annotation durations for the BoundaryNet-based approach are much smaller compared to the other approaches, despite BoundaryNet being a semi-automatic approach. This is due to the  superior quality contours generated by BoundaryNet which minimize post-inference manual correction burden.

\section{Conclusion}

In this paper, we propose BoundaryNet, a novel architecture for semi-automatic layout annotation. The advantages of our method include (i) the ability to process variable dimension input images (ii) accommodating large variation in aspect ratio without affecting performance (iii) adaptive boundary estimate refinement. We demonstrate the efficacy of BoundaryNet on a diverse and challenging document image dataset where it outperforms competitive baselines. Finally, we show that BoundaryNet readily generalizes to a variety of other historical document datasets containing dense and uneven layouts. Going ahead, we plan to explore the possibility of incorporating BoundaryNet into popular instance segmentation frameworks in an end-to-end manner. 

\bibliographystyle{splncs04}
\bibliography{main}

\end{document}